# CAUSAL DEEP Q NETWORK

Elouanes Khelifi[1] and Amir Saki[2] and Usef Faghihi[3]

[1] Université du Québec à Trois-Rivières, Québec G8Z 4M3, Canada

[1] `Elouanes.khleifi@uqtr.ca`
[2] `Amir.saki.math@gmail.com`
[3] `Usef.Faghihi@uqtr.ca`

**Abstract.** Deep Q Networks (DQN) have shown remarkable success in various reinforcement learning tasks. However, their reliance on associative learning often leads to the acquisition of spurious correlations, hindering their problem-solving capabilities. In this paper, we introduce a novel approach to integrate causal principles into DQNs, leveraging the PEACE (Probabilistic Easy vAriational Causal Effect) formula for estimating causal effects. By incorporating causal reasoning during training, our proposed framework enhances the DQN's understanding of the underlying causal structure of the environment, thereby mitigating the influence of confounding factors and spurious correlations. We demonstrate that integrating DQNs with causal capabilities significantly enhances their problem-solving capabilities without compromising performance. Experimental results on standard benchmark environments showcase that our approach outperforms conventional DQNs, highlighting the effectiveness of causal reasoning in reinforcement learning. Overall, our work presents a promising avenue for advancing the capabilities of deep reinforcement learning agents through principled causal inference. The code is available on GitHub[1].

## 1    Introduction

Deep Q Networks (DQNs) have proven to be effective tools for tackling complex decision-making tasks in reinforcement learning (RL). By accurately estimating the optimal action-value function, DQNs equip agents with the ability to learn and apply successful strategies in various environments [1]. However, their effectiveness is sometimes constrained by a significant challenge: the dependence on associative learning mechanisms [2]. This reliance may result in the development of spurious correlations and erroneous decision-making. These spurious correlations can arise due to confounding variables or non-causal relationships present in the data, resulting in suboptimal performance and limited generalization capabilities [2, 3].

To address this challenge, recent research has focused on integrating causal reasoning into DQNs [4-6], aiming to enhance their understanding of the underlying causal structure of the environment. By incorporating causal principles, such as identifying and adjusting for confounding variables, DQNs (and other reinforcement learning architectures) can potentially improve their decision-making accuracy and robustness [7]. In this paper, we propose a novel approach to infuse causal reasoning into DQNs, introducing the PEACE (Probabilistic Easy vAriational Causal Effect ) [8] formula for estimating causal effects.

Our primary motivation stems from the observation that traditional DQNs lack the ability to discern between actions that cause rewards and mere correlations caused by confounding factors, this leads to suboptimal performance in complex environments where causal understanding plays a crucial role as we show the results section [8]. The PEACE formula allows DQNs to disentangle causal relationships from spurious correlations during training. By leveraging PEACE, we aim to alleviate the limitations of traditional DQNs and enhance their problem-solving capabilities without sacrificing performance.

In the upcoming sections, we delve into the latest advancements in related fields, lay out the theoretical groundwork for our method, and explain how we incorporate the PEACE formula into DQNs' training regimen. We also assess its effectiveness across standard testing environments. Our findings reveal that enhancing DQNs with causal reasoning capabilities markedly boosts their capacity to discern causal connections, which in turn leads to improved decision-making and outperforms traditional DQNs. Our research marks a notable leap forward in deep reinforcement learning by merging associative learning with causal analysis, setting the stage for the development of more reliable and understandable RL agents.

---

[1] https://github.com/joseffaghihi/CausalRL.git

In this paper, we assume the RL framework follows the causal structure described below (see Figure 1). Here, E represents the environment of the agent according to the Reinforcement Learning framework's definition [9]. The actions taken by the agent are represented by the node A. The rewards returned by the RL environment are represented by the node R, and Z represents a hidden confounder which causes spurious correlations. According to our causal structure, the confounder Z influences both the environment and the rewards and causes it to put the agent in certain states that are likely to cause the agent to choose certain actions. These actions will result in sub-optimal rewards which in turn, encourage the agent to perform more of them.

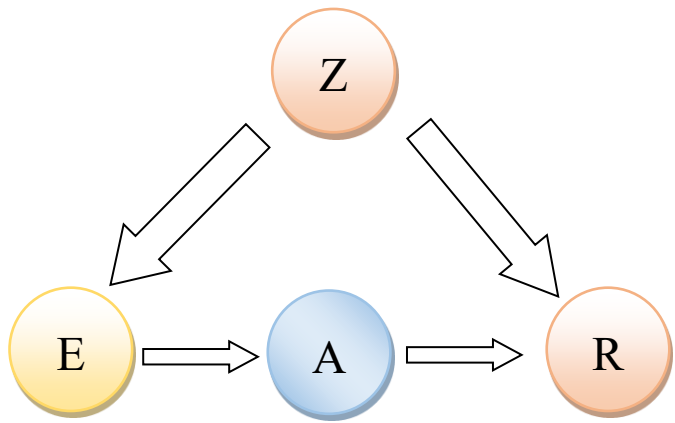

**Fig. 1.** This Directed Acyclic Graph (DAG) represents the causal structure of the RL framework entities. Where Z is the set of hidden confounding variables. A is the action variable. R is the reward variable, and E is the RL environment, which is a function G(S,A) = (R,S'), where R is the reward, S is the current state, S' is the next state and A is the action. The directed arcs represent the causal effect on one variable to the other.

The confounder Z causes the agent to learn a sub-optimal policy entailed by the spurious correlation between action and rewards. In an ideal setup, there should not be any factor that causes environment to give out more rewards so that the agent learns the true dynamics of the environment, and the optimal policy that truly causes the rewards. In our work we attempt to solve this by eliminating the confounding variables and forcing the agent to learn a policy that enhances the probability of taking actions that are causally related to rewards.

## 2    Related Works

In the quest to integrate causality with reinforcement learning, the study by CHH. Yang and colleagues presents a notable development known as the Causal Inference Q-network (CIQ) [4]. The CIQ framework is designed to tackle the key issue of observation interference, which can arise from various factors such as noise, adversarial attacks, black-outs, or frozen frames. During training, the framework incorporates an auxiliary interference label, and during testing, it employs a neural network architecture tailored to estimate the condition of interference. This approach enables the agent to make decisions by considering the estimated state of interference, thereby concentrating on pertinent information despite disruptive elements. Consequently, the CIQ framework generates Q-values that direct the agent towards actions that optimize rewards, enhancing the resilience of reinforcement learning to external perturbations. Nonetheless, the CIQ framework also brings additional layers of complexity to the conventional RL structure. It necessitates the estimation of interference labels and the implementation of a mechanism to switch between neural networks, which may result in greater computational demands throughout the training and inference phases.

A. Méndez et al. [5] present another significant approach to address the issue of spurious association in policy learning within reinforcement learning (RL). Their paper proposes an enhancement of RL efficiency by incorporating causal relationships between state variables and actions. The core concept revolves around simplifying the state or action space by employing causal models to streamline the agent's decision-making process. The underlying assumption is that causal inference can bolster RL through the exploitation of causal links between state variables, actions, and objectives, potentially reducing the dimensionality of the state or action space and thereby facilitating more efficient learning. However, it is critical to recognize that real-world causal relationships often exhibit a high degree of complexity and nonlinearity. The effectiveness of the Méndez method hinges on the precision and comprehensiveness of the causal models it utilizes. In intricate environments, constructing accurate causal models poses a considerable challenge.

Q-Cogni framework [10] aims to combine the strengths of model-free and model-based reinforcement learning approaches by incorporating causal structure discovery. The Q-Cogni framework utilizes autonomous causal structure discovery to help the reinforcement learning agent deal with imperfect environments, such as latent variables, while maintaining sample efficiency. The method uses the NOTEARS algorithm to learn the causal structure and Bayesian Networks to make inferences, with the possibility of incorporating expert domain knowledge to improve the model. The approach could potentially face limitations in terms of scalability and computational demand when applied to very complex or dynamic environments.

*Yu et al.* [11] framework constructs a causal world model that captures the dynamics of the environment. It uses causal discovery to identify the causal graph and fit causal equations for the variables, allowing for causal chain analysis and explanations about the agent's decisions. This approach is promising for generating explainable reinforcement learning models, which can be particularly useful for interpretability and trust in AI systems. The framework, however, may struggle with efficiency when dividing data to learn structural causal models (SCMs) and could also encounter challenges with redundant parameters and reduced sample efficiency. Moreover, adapting to new causal relationships discovered through the agent's exploratory behaviors adds another layer of complexity.

It's important to highlight that the model presented in [11] is considerably more complex and not directly comparable to our more mathematically concise approach. Furthermore, the absence of publicly accessible code for the studies [4, 5, 10] prevented a direct comparison of our findings with theirs.

In our approach, we attempt to solve the complexity issues by using the PEACE formula, which allows for the integration with backpropagation during training. This equips our framework with causal reasoning capabilities and enhances the performance of our agent without adding the additional complexity associated with the models mentioned above.

## 3 Causal DQN (C-DQN)

Before delving into the details of our Causal DQN model and its application in enhancing Deep Q Networks (DQNs), it is essential to understand its theoretical underpinnings and significance in causal inference. The PEACE formula [8] represents a derivable and precise causal model, offering a principled framework for estimating causal effects in the presence of confounding variables. At its core, the PEACE formula leverages probabilistic variational techniques to quantify the interventional changes in the outcome variable Y attributable to a specific variable of interest X, while accounting for the influence of other variables encapsulated in Z. By systematically evaluating the probabilistic variations in Y as X transitions through its possible values, the PEACE formula provides a rigorous methodology for disentangling causal relationships from mere associations. This introduction sets the stage for a detailed exploration of the PEACE formula and its integration into the training process of DQNs, thereby enhancing their ability to learn causal dependencies and make informed decisions in reinforcement learning tasks.

## 3.1 PEACE

Let's consider that $Y = g(X, Z)$, where $X$ is a finite random variable that we are interested in its causal effect on $Y$, and $Z$ is the random vector of all other variables directly affecting $Y$. By $g_{in}(x, z)$, we mean to replace all equations describing $X$ and $Z$ in the model by $X = x$ and $Z = z$ (i.e., intervening on the values of $X$ and $Z$). Following [8], for each specific value $Z = z$, we define the probabilistic interventional easy variation of $X$ on $Y$ as follows:

$$PIEV^z(X \to Y) = \sum_{i=0}^{l} |g_{in}(x_i, z) - g_{in}(x_{i-1}, z)| P(x_i | z) P(x_{i-1} | z) \quad (1)$$

where we have assumed that $x_0, \ldots, x_l$ are all possible values of $X$ with $x_0 < \cdots < x_l$. Intuitively, we start from the smallest value of $X$ and gradually increase the value of $X$ while calculating the interventional changes of $Y$. Here, $P(x_i | z) P(x_{i-1} | z)$ is the probability of independently selecting $x_{i-1}$ and $x_i$ given $z$, and it captures how the interventional change from $x_{i-1}$ to $x_i$ given $z$ is available. Finally, the PEACE of $X$ on $Y$ is defined to be the expected value of $PIEV^z(X \to Y)$ with respect to $Z$, namely:

$$PEACE(X \to Y) = E_Z(PIEV^z(X \to Y)) \quad (2)$$

In practice, $Z$ consists of two parts, one is observed, and another one is not. Another issue is that for each unit under the experiment just one of the values of $X$ given $Z = z$ could be observed, and hence we suffer from the fundamental problem of causal inference. To deal with these two issues, we need to identify the PEACE formula. To do so, we use some assumptions as described in the following theorems. We say that a function $h(x, z, u)$ is separable with respect to $z$ whenever there exist two functions $h^{(1)}(x, z)$ and $h^{(2)}(z, u)$ with:

$$h(x, y, z) = h^{(1)}(x, z) + h^{(2)}(z, u) \quad (3)$$

**Theorem**. Assume that in the following $Z$ denotes the observed covariates while $U_Y$ denotes the unobserved one. By $Y_{x,z}$ we mean intervening by $X = x$ and $Z = z$ (i.e., replacing the equations describing the variables $X$ and $Z$ by $X = x$ and $Z = z$, respectively).

1. $g_{in}(X, Z, U_Y)$ is separable with respect to $Z$, and
2. $X$ and $Z$ are independent given $U_Y$.

Then, $PEACE(X \to Y)$ is identifiable. Further, assume that we have the following extra assumptions for any $X = x$ and $Z = z$:

3. $Y_{x,z}$ and $X$ are independent given $Z$,
4. $Y_{x,z}$ is a one-to-one function of $(U_Y)_{x,z}$, and
5. $Y_{x,z}$ and $Z$ are independent.

Then, we have that:

$$PIEV^z(X \to Y) = \sum_{i=1}^{l} |E(Y | x_i, z) - E(Y | x_{i-1}, z)| P(x_i | z) P(x_{i-1} | z). \quad (4)$$

We note in the above theorem, when we involve all important variables, we will have something like $Y = g(X, Z) + U_Y$, and $U_Y$ is unrelated to $X$ and $Z$. Hence, Conditions 1, 2 and 4 are satisfied, and the two other assumptions are some versions of ignorability and conditional ignorability.

## 3.2 C-DQN Architecture

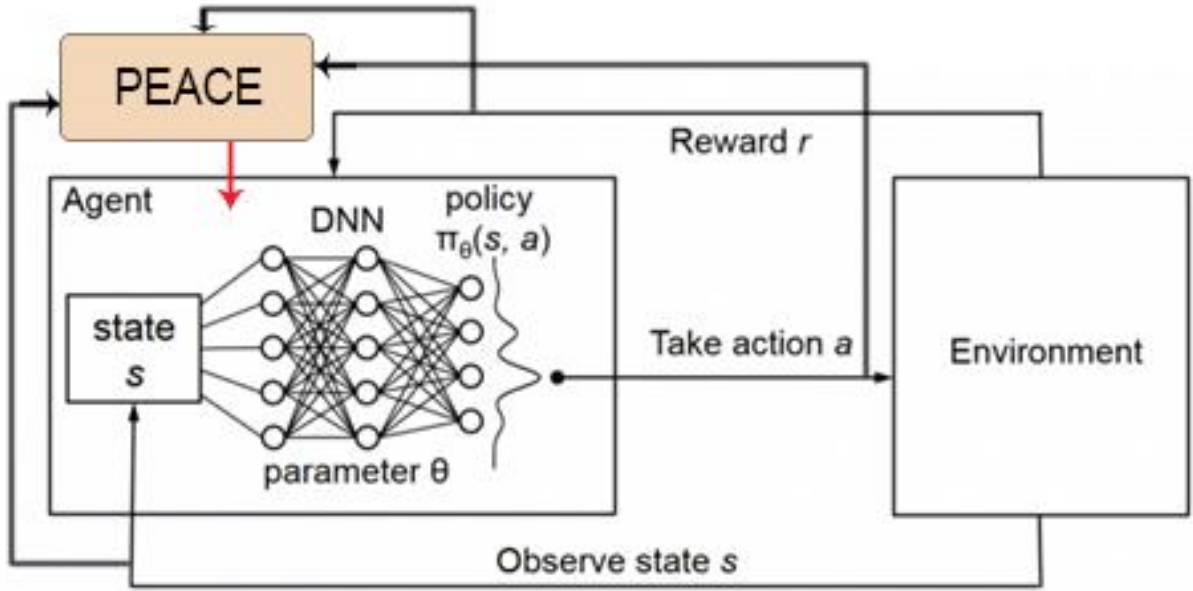

**Fig. 2.** Causal DQN Pipeline, the redline represents the injection of causal effect estimate into the loss of the neural network of the DQN Agent (Image modified from [12])

The core idea in our model is using the PEACE formula to add causal reasoning capabilities to the DQN model (and possibly other RL models). To do so, we use the PEACE formula above to eliminate the hidden confounder Z (see Figure 1) and estimate the true causal effect between the actions taken and the rewards. We sample from the minibatch used in the training pipeline and compute PEACE of each minibatch. The estimated causal effect is then added as a penalty term as in equation (5). The value of the penalty term decreases when the causal effect between action and reward is higher, and through backpropagation, this encourages the network to choses actions that are more causally related to the rewards.

$$\mathcal{L} = r + (\gamma \max Q(S, a') - Q(S, a))^2 + \frac{1}{PEACE(a,r)} \quad (5)$$

After several iterations the penalty (as shown in the result section) shifts the weights of the policy network towards values that prioritize actions that are more causally related to rewards. The agent then acquires a stronger understanding backed by a causal reasoning strategy instead of merely associating actions to rewards in a purely statistical manner.

# 4    Results

In this section we will compare the performance of DQNs equipped with the PEACE formula against traditional DQNs.

To assess the efficacy of our proposed method, we conducted comprehensive tests using standard reinforcement learning benchmarks, including the OpenAI Gym library [13]. This toolkit facilitates the development and comparison of reinforcement learning algorithms by providing a diverse array of standardized environments. These range from straightforward toy problems to more sophisticated challenges, such as simulated robotic control tasks, thereby offering a robust platform for evaluating RL models.

We assessed the agents' ability to learn optimal policies in environments with varying degrees of complexity, including both deterministic and stochastic settings. Additionally, we examined the agent's sensitivity to spurious correlations and their generalization capabilities across unseen scenarios. Our results demonstrate that integrating causal principles into DQNs leads to substantial improvements in problem-solving capabilities, with negligeable impact on performance metrics such as convergence speed and computational efficiency. We present our findings in terms of both quantitative performance metrics and qualitative analysis, highlighting the advantages of our approach in mitigating spurious correlations and facilitating more robust decision-making in challenging environments.

In Figure 3 and 4, we compare the performance of a conventional Deep Q Network (DQN) with our proposed Causal DQN in the classic CartPole environment provided by the OpenAI Gym toolkit [13]. The CartPole environment serves as a standard benchmark for evaluating discrete action space reinforcement learning algorithms, requiring agents to balance a pole on a cart by applying appropriate actions. We trained both the conventional DQN and the Causal DQN on the same configuration and hyperparameters to ensure a fair comparison.

Our results indicate that the Causal DQN consistently outperforms the conventional DQN across various performance metrics. Specifically, the Causal DQN exhibits improved stability in learning the optimal policy, as evidenced by faster convergence and higher average in episode rewards.

During training, we tracked episode rewards, average test rewards, and the total causal effect using the PEACE formula. We implemented an early stopping criterion for both agents at a score threshold of 200 [14], ceasing training once the agent's average reward met the solved environment's benchmark. This method underscores the effectiveness of the PEACE-equipped agents. Figure 4 shows the Causal agent solving the cartpole environment in 147 episodes, about 3.6 times faster than the standard DQN, with higher average rewards and more stable learning progression.

Our results demonstrate that our causal agent identifies actions more causally linked to rewards compared to the traditional DQN agent. The Causal DQN shows a better grasp of the causal dynamics within the CartPole environment. Utilizing the PEACE formula, the Causal DQN effectively distinguishes between causal and non-causal correlations, enhancing decision-making accuracy and interpretability.

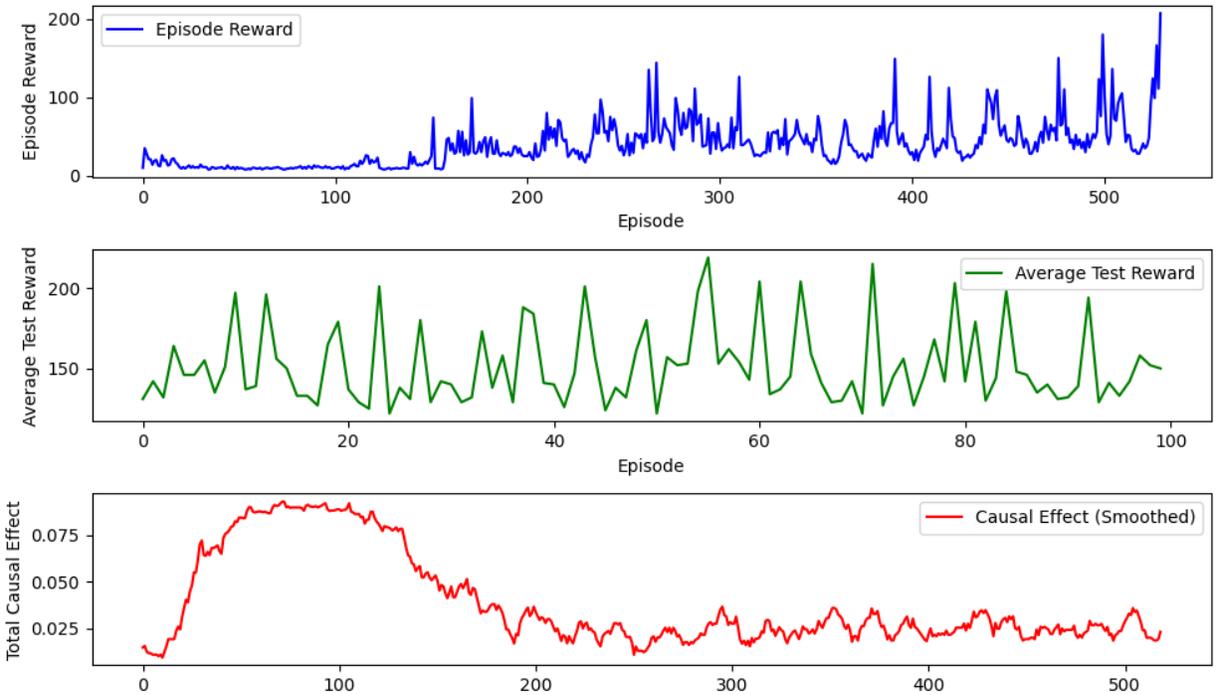

**Fig. 3.** Normal DQN benchmarks after 530 episodes, the top plot shows the rewards during training. The middle plot shows the rewards during the testing phase, and the bottom graph shows the average total causal effect during training for each episode.

Next, we pitted the two trained agents, in Figure 5, dubbed *'Agent Smith'*, for the normal DQN and *'Causal-Smith'* for the causal DQN, against each other in the Cartpole environment for 100 rounds, with each round consisting of 5 episodes. The average score per round, derived from these 5 attempts, is depicted in Figure 5. The causal agent, on average, scored 350, significantly outperforming the normal DQN's average score of 120. This substantial performance boost, amounting to a 290% increase, is attributed to the causal agent's superior understanding of the task's causal structure, as indicated by its higher total causal effect.

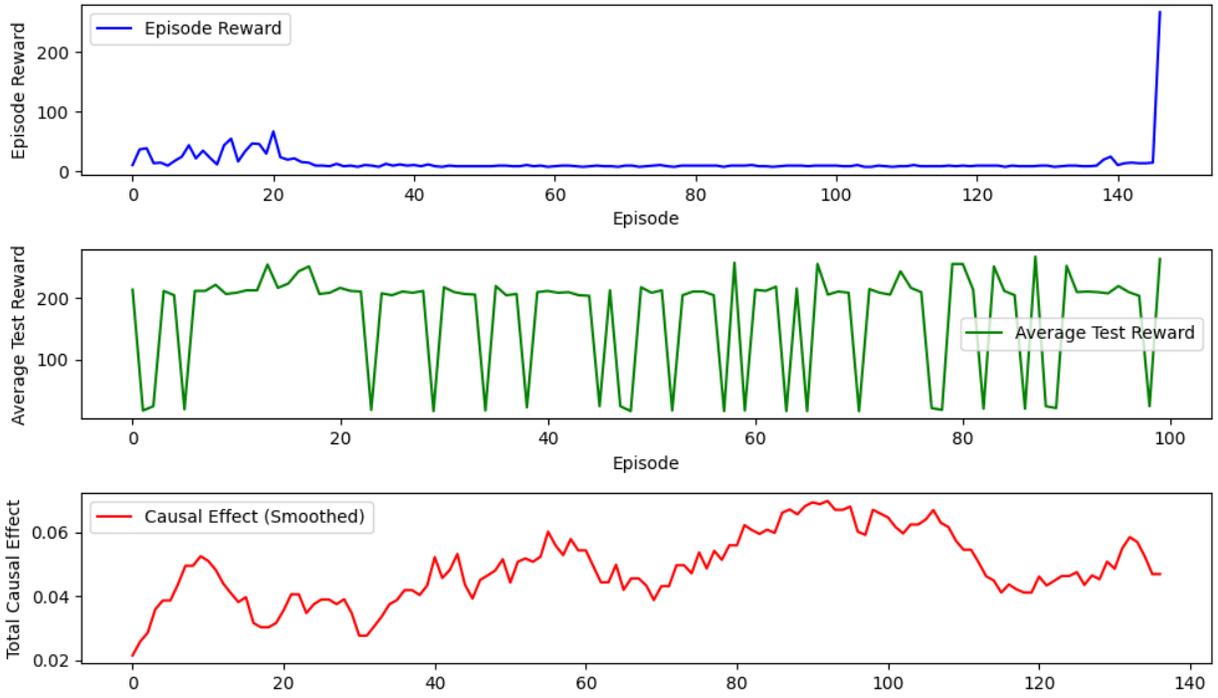

**Fig. 4.** Causal DQN benchmarks after 147 episodes (same metrics as Figure 1).

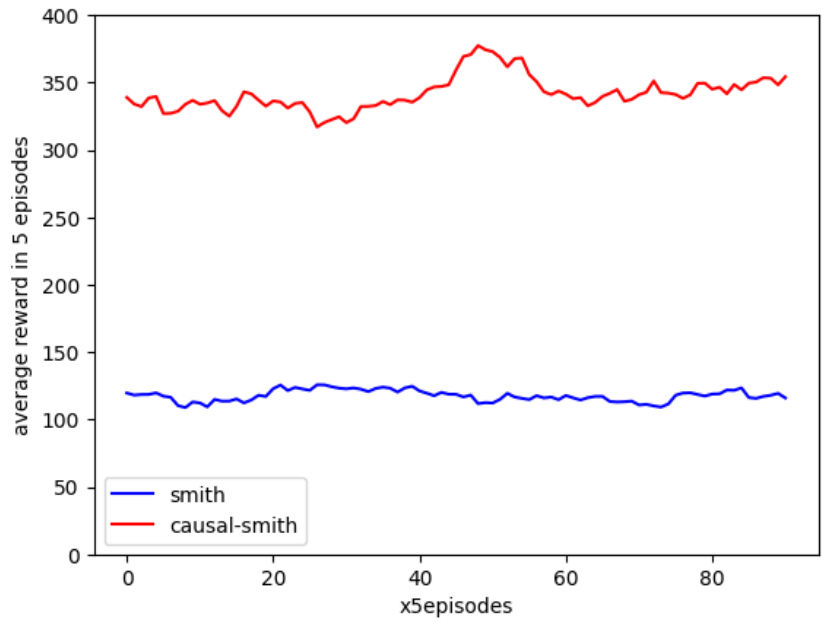

**Fig. 5.** Causal DQN vs regular DQN

Our findings underscore the effectiveness of incorporating causal principles into DQNs, with the Causal DQN outperforming its counterpart significantly. This success underlines the PEACE causal reasoning model's potential to improve deep reinforcement learning agents' proficiency in complex decision-making scenarios.

## 5    Conclusion and Future Works

This work illustrates the significant advantages of integrating causal reasoning, specifically through the PEACE formula, into Deep Q Networks (DQNs) for reinforcement learning. By addressing the limitations of traditional DQNs, which struggle with spurious correlations, our approach enhances decision-making accuracy and robustness. The empirical evidence demonstrated through rigorous testing in environments like the OpenAI Gym's CartPole, confirms our causal agent's superior performance. This research not only advances deep reinforcement learning by blending causal analysis with associative learning but also sets a new precedent for developing more intelligent, reliable, and interpretable reinforcement learning agents.

In future work, we plan to extend our framework to accommodate reinforcement learning problems with continuous action spaces. This involves using a modified version of the PEACE formula to work with continuous treatment values, enabling the integration of causal reasoning into Deep reinforcement learning models such as A3C and PPO for tasks requiring continuous control. Additionally, we aim to assess the scalability and generalizability of our approach across diverse reinforcement learning domains beyond the virtual environments. We will also focus on enhancing the interpretability of Causal Reinforcement Learning models and exploring techniques for transferring causal knowledge between tasks to improve learning efficiency. Overall, our future research aims to advance the capabilities of Causal inference in Reinforcement Learning and establish causal reasoning as a core component of this sub-field of machine learning.

**Acknowledgments.** This study was funded by Kobotik inc[2] and Mitacs[3]

---